\documentclass[letterpaper]{article} 
\usepackage{aaai25}  
\usepackage{times}  
\usepackage{helvet}  
\usepackage{courier}  
\usepackage[hyphens]{url}  
\usepackage{graphicx} 
\usepackage{natbib}  
\usepackage{caption} 
\usepackage{algorithm}
\usepackage{algorithmic}

\usepackage{newfloat}
\usepackage{listings}
\usepackage{comment}
\usepackage{color}

\usepackage{xurl}

\usepackage{tabularx}

\usepackage{tikz}
\usetikzlibrary{arrows.meta, positioning, shapes.geometric}
\usetikzlibrary{fit, backgrounds, calc}

\usetikzlibrary{positioning, shapes.geometric, arrows.meta, fit}
\tikzstyle{phase} = [rectangle, draw=black, thick, rounded corners=2pt, text width=6cm, minimum height=1.2cm, align=center, fill=blue!10]
\tikzstyle{substep} = [rectangle, draw=gray, thick, rounded corners=2pt, text width=5.5cm, minimum height=1cm, align=center, fill=gray!10, font=\small]
\tikzstyle{arrow} = [thick,->,>=Stealth]

\DeclareCaptionStyle{ruled}{labelfont=normalfont,labelsep=colon,strut=off} 
\floatstyle{ruled}
\newfloat{listing}{tb}{lst}{}
\floatname{listing}{Listing}
\title{Towards Developing Standards and Guidelines for Robot 
Grasping\\and Manipulation Pipelines in the COMPARE Ecosystem}
\author{
    Huajing Zhao,
    Brian Flynn, 
    Adam Norton,
    Holly Yanco
}
\affiliations{
    New England Robotics Validation and Experimentation (NERVE) Center\\
    University of Massachusetts Lowell, Lowell, MA, USA\\

    huajing\_zhao@uml.edu, brian\_flynn@uml.edu, adam\_norton@uml.edu, holly\_yanco@uml.edu
}

\begin{document}

\maketitle

\begin{abstract}
The COMPARE Ecosystem aims to improve the compatibility and benchmarking of open-source products for robot manipulation through a series of activities.
One such activity is the development of standards and guidelines to specify modularization practices at the component-level for individual modules (e.g., perception, grasp planning, motion planning) and integrations of components that form robot manipulation capabilities at the pipeline-level.
This paper briefly reviews our work-in-progress to date to (1) build repositories of open-source products to identify common characteristics of each component in the pipeline, (2) investigate existing modular pipelines to glean best practices, and (3) develop new modular pipelines that advance prior work while abiding by the proposed standards and guidelines.
\end{abstract}

\section{Introduction}

Executing grasping and manipulation tasks with a robot system requires the integration of many hardware and software subsystems. 
Most researchers generally only contribute one novel subsystem at a time and the other subsystems used are generally not standardized.  Each research group often crafts their own custom software stack (often utilizing ROS~\cite{quigley2009ros}), making it difficult to conduct true side-by-side comparisons. 
The Collaborative Open-source Manipulation Performance Assessment for Robotics Enhancement (COMPARE) Ecosystem aims to create a greater cohesion between open-source products towards improving the modularity of software components in the robot manipulation pipeline for more effective performance benchmarking\footnote{https://www.robot-manipulation.org/}. 
After leading a series of workshops and surveys for the robot manipulation community, several limitations of the current ecosystem with recommendations for improvement were determined, including issues related to integrating open-source products and a lack of truly modular software to enable both component-level and holistic system evaluations.
These recommendations align with previously published sentiments around issues with code reproducibility in robotics~\cite{cervera2023run,cervera2018try}.
Recommendations for improvement include organized repositories of open-source products and benchmarking assets, and the benefits of modular software components was highlighted.

With these recommendations in mind, development of the COMPARE Ecosystem includes the production of standards and guidelines for topics including developing open-source products, conducting benchmarking evaluations, and reporting performance results. 
Test and evaluation procedures will be developed to validate the open-source products and their compatibility with others in the software pipeline, including reviews of their adherence to the developed standards and guidelines, performance assessments of their functionality, and usability analyses when attempting to integrate the products. 
To this end, we are developing standards and guidelines for improving the modularity and interoperability of software components in the manipulation pipeline with an initial focus on object grasping and pick-and-place capabilities.
This paper presents our work-in-progress to date.

\section{Scope}

We consider open-source software at two levels: \textit{components} that drive the underlying functions of a manipulation capability (e.g., scene segmentation, pose estimation, grasp planning, motion planning, execution) and \textit{pipelines} that integrate multiple components together to perform a task (e.g., pick and place, assembly, disassembly, door opening). 
\textbf{Component-level} guidelines will include recommendations for specifying dependencies and resolving compatibility issues that can apply to many types of open-source software contributions.
\textbf{Pipeline-level} standards will specify common interfacing techniques between components including input/output data formats and ROS service architectures, to include execution of robot capabilities (e.g., object grasping) and benchmarking procedures (e.g., lifting and shaking the object to demonstrate a firm grasp), such that components can be easily swapped in and out (i.e., making them modular and interoperable).
The goal is to lower the barrier to integration of open-source software which in turn will improve reproducibility of robot capabilities between labs.

Our initial use case is picking in clutter for which there are many types of open-source products and benchmarking assets available.
While other use cases in robot manipulation will also benefit from these efforts (e.g., assembly, cloth manipulation), the capabilities and evaluations for picking in clutter are often leveraged in these applications, making for a good starting point.
The scope of this effort will expand as it matures and the community grows.

\section{Establishing Standards and Guidelines}

Three efforts are underway to establish these standards and guidelines for the COMPARE Ecosystem: (1) building repositories of open-source software components, (2) investigating prior efforts to produce modular pipelines, and (3) using lessons learned to produce new pipelines that utilize the proposed standards and guidelines.

\subsection{Building Repositories of Open-Source Products}

\textit{Robot-Manipulation.org} -- the website for the COMPARE Ecosystem -- hosts several repositories of available open-source products and benchmarking assets.
These repositories were initially developed through literature reviews into prominent contributions for each type of relevant software component.
The gathered resources were then characterized according to a set of parameters that would (a) be useful for a researcher looking for a component that matches their needs and (b) to aid in establishing how each component functions in a pipeline such that commonalities could be identified.
For example, the \textit{Grasp Planning} category\footnote{https://www.robot-manipulation.org/software/grasp-planning} consists of repositories for Grasp Planners (e.g., EquiGraspFlow~\cite{lim2024equigraspflow}) and  Grasp Datasets (e.g., GraspClutter6D~\cite{back2025graspclutter6d}).
The organizational axes of Grasp Planners include planning method (e.g., sampling, direct regression, reinforcement learning), end-effector hardware (e.g., suction, two-finger, multi-finger), input data (e.g., point cloud, depth image, RGB image), and output pose format (e.g., 2D grasp rectangle, 6-DoF grasp, 7-DoF grasp).  
Grasp Datasets includes similar axes as well as metrics like number of grasps and objects included, among other parameters.

Additional repositories for open-source software components include: \textit{Motion Planning} with Motion Planners and Motion Planning Datasets; \textit{Perception} with Scene Segmentation, Object Detection, and Pose Estimation; \textit{Learning} with Learning Environments; and \textit{Simulation} with Simulators and Models and Descriptions.
By gathering this information, the COMPARE project team can determine the various classes of standards and guidelines that will be developed based on shared salient characteristics.
Using the \textit{Grasp Planning} example, input data for a planner will impact the relevant perception components and how their output data is shared, while end-effector hardware will impact output pose format; this info can be used to form the basis for \textbf{pipeline-level} standards.
The repositories are continuously updated as new research is published and new open-source products are released, either researched by the COMPARE project team or submitted by the community.

\subsection{Investigating Existing Modular Pipelines}

Towards designing common standards and guidelines, we can leverage existing open-source pipelines with modular components that can be swapped out for evaluation purposes
For example, the GRASPA project~\cite{bottarel2023graspa} specifies a software framework with a grasp planning benchmarking protocol provided in a reproducible Docker container and three grasp planners to use.
Another example is SceneReplica~\cite{khargonkar2024scenereplica}, which specifies a tool to aid in reproducing multi-object scenes for benchmarking pick-and-place tasks, but also includes pipelines of modular, interchangeable components (perception, grasp planning, motion planning, and control) that are used to produce a spread of performance data using the benchmark.
Two perception-to-action pipelines are provided -- \textit{model-based grasping} (e.g., planning precomputed grasps of known objects) and \textit{model-free grasping} (e.g., planning grasps of unknown objects) -- developed in ROS  1 with fully open-source simulation and evaluation functionality that leverages multiple existing open-source components. 

Due to the pipelines provided by SceneReplica supporting multiple types of interchangeable components -- rather than just grasp planning --  we opted to investigate it further.
The model-based solution provides image-based object pose estimation (e.g., PoseCNN~\cite{xiang2017posecnn}, PoseRBPF~\cite{deng2021poserbpf}) and offline grasp optimization (e.g., GraspIt!~\cite{miller2004graspit})  models pretrained on known objects such as those from the YCB Object and Model Set~\cite{calli2015ycb}. 
The model-free solution provides online object segmentation (e.g. Unseen Object Clustering (UOIS)~\cite{xiang2020learning}, MSMFormer~\cite{https://doi.org/10.48550/arxiv.2211.11679}) and point-cloud-based grasp detection (e.g., 6-DoF GraspNet~\cite{mousavian20196}, Contact-GraspNet~\cite{sundermeyer2021contact}) models on unseen objects, enabling category-agnostic manipulation through raw RGB-D input.

We first attempted to replicate the functionality in our own lab from their source code\footnote{https://github.com/IRVLUTD/SceneReplica}, implementing both pipelines with modular components within a Docker container environment with ROS  1. 
Early experiments involved running modular segments both independently and integrated within the perception-to-action pipelines, as well as validating GPU usage isolation to diagnose and optimize memory bottlenecks.
We resolved multiple compatibility issues involving CUDA-based computing~\cite{luebke2008cuda} and graphic rendering~\cite{eng2019qt5, hill2000computer, schroeder2000visualizing} inside a Docker container, establishing a stable runtime configuration that can be reproduced across systems. 
Through this exercise, we are extracting best practices for integration to inform future \textbf{component-level} guidelines that will support standardized deployment and direct comparisons between pipelines with varying dependencies.

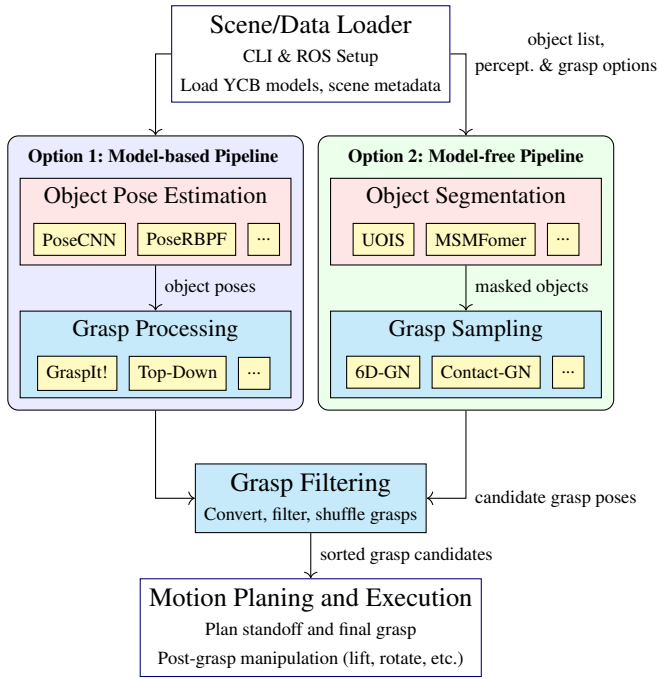
\begin{figure}[htp]
    \centering
\begin{tikzpicture}[node distance=6mm and -12mm, every node/.style={align=center}]
\node[draw, rectangle, draw=blue!50!black] (loader) { Scene/Data Loader\\ \scriptsize CLI \& ROS Setup\\ \scriptsize Load YCB models, scene metadata};
\node[draw, rectangle, minimum width=36mm, below left= 10mm and -16mm of loader, fill=pink!40,font=\small] (pose) { Object Pose Estimation \\   \\};
\node[draw, rectangle, minimum height=4.25mm, below left= 16mm and 6mm of loader, fill=yellow!30] (poseCNN) {\scriptsize PoseCNN};
\node[draw, rectangle, minimum height=4.25mm, right= 1.5mm of poseCNN, fill=yellow!30] (poseRBPF) {\scriptsize PoseRBPF};
\node[draw, rectangle, minimum height=4.25mm, right=1.5mm of poseRBPF, fill=yellow!30] (poseTBA) {\scriptsize ...};
\node[draw, rectangle, minimum width=36mm,  below= of pose, fill=cyan!20 ,font=\small] (grasp_preproc) { Grasp Processing\\  \\ };
\node[draw, rectangle, minimum height=4.25mm, below left= 12mm and -13mm of pose, fill=yellow!30] (graspit) {\scriptsize GraspIt!};
\node[draw, rectangle, minimum height=4.25mm, right=1.5mm of graspit, fill=yellow!30] (grasp_td) {\scriptsize Top-Down};
\node[draw, rectangle, minimum height=4.25mm, right=1.5mm of grasp_td, fill=yellow!30] (graspTBA) {\scriptsize ...};
\node[draw, rectangle,  minimum width=36mm, below right= 10mm and -16mm of loader, fill=pink!40,font=\small] (segproc) { Object Segmentation\\  \\ };
\node[draw, rectangle, minimum height=4.25mm, below right= 16mm and -13mm of loader, fill=yellow!30] (uois) {\scriptsize UOIS};
\node[draw, rectangle, minimum height=4.25mm, right= 1.5mm of uois, fill=yellow!30] (msm) {\scriptsize MSMFomer};
\node[draw, rectangle, minimum height=4.25mm, right=1.5mm of msm, fill=yellow!30] (segTBA) {\scriptsize ...};
\node[draw, rectangle, minimum width=36mm, below=of segproc, fill=cyan!20,font=\small] (graspnet) {Grasp Sampling \\ \\};
\node[draw, rectangle, minimum height=4.25mm, below left= 12mm and -12mm of segproc, fill=yellow!30] (6dof_graspnet) { \scriptsize 6D-GN};
\node[draw, rectangle, minimum height=4.25mm, right= 1.5mm of 6dof_graspnet, fill=yellow!30] (contact_graspnet) { \scriptsize Contact-GN};
\node[draw, rectangle, minimum height=4.25mm, right= 1.5mm of contact_graspnet, fill=yellow!30] (graspTBA2) { \scriptsize ...};
\node[draw, rectangle, below=48mm of loader, fill=cyan!20] (filter) { Grasp Filtering\\ \scriptsize Convert, filter, shuffle grasps};
\node[draw, rectangle, below= of filter, draw=blue!50!black] (planning) {Motion Planing and Execution \\ \scriptsize Plan standoff and final grasp \\ \scriptsize Post-grasp manipulation (lift, rotate, etc.)
};
\begin{scope}[on background layer]         
  \node[draw=none,
        fill=none,           
        inner sep=1.5mm,          
        fit=(pose) (grasp_preproc)] (MB) {} ; 
\node[coordinate] (dummyT) at ($(MB.north)+(0,4mm)$) {};
\end{scope}
\begin{scope}[on background layer]
  \node[draw=black,
        fill=blue!8,
        rounded corners,
        inner sep=0pt,         
        fit=(MB)(dummyT)] (MB_shader) {};  

  \node[anchor=north,         
        font=\scriptsize\bfseries, 
        align=center,
        xshift=0mm, yshift=5mm]   
        at (pose.north)      
        {\textbf{Option 1: Model-based Pipeline }}; 
\end{scope}
\begin{scope}[on background layer]          
  \node[draw=none,
        fill=none,           
        inner sep=1.5mm,          
        fit=(segproc) (graspnet)] (MF) {} ; 
\node[coordinate] (dummyT) at ($(MF.north)+(0,4mm)$) {};
\end{scope}
\begin{scope}[on background layer]
  \node[draw=black,
        fill=green!8,
        rounded corners,
        inner sep=0pt,         
        fit=(MF)(dummyT)] (MF_shader) {};  
  \node[anchor=north,         
        font=\scriptsize\bfseries, 
        align=center,
        xshift=0mm, yshift=5mm]   
        at (segproc.north)      
        {\textbf{Option 2: Model-free Pipeline}}; 
\end{scope}
\draw[->] (loader) -| (MB_shader);
\draw[->] (loader) -| node[right, font=\small]{\scriptsize object list, \\ \scriptsize percept. \& grasp options} (MF_shader);
\draw[->] (pose) -- node[right, font=\small]{\scriptsize object poses} (grasp_preproc);
\draw[->] (MB_shader) |- (filter);
\draw[->] (segproc) -- node[right, font=\small]{\scriptsize masked objects} (graspnet);
\draw[->] (MF_shader) |- node[right, font=\small]{\scriptsize candidate grasp poses} (filter);
\draw[->] (filter) --  node[right, font=\small]{\scriptsize sorted grasp candidates}(planning);
\end{tikzpicture}
    \caption{Block diagram for perception-to-action pipelines in SceneReplica using ROS  1 with interchangeable model-based and model-free grasp planning options.}
    \label{fig:sceneRep_flow}
\end{figure}

Next, we further studied the source code and consulted with the authors for a detailed understanding of how the pipelines functioned; see Fig.~\ref{fig:sceneRep_flow} for a high-level block diagram of the pipelines' construction and data that is passed between stages. 
The diagram in Fig.~\ref{fig:sceneRep_flow} reflects our interpretation of the workflow after successfully implementing the pipelines of SceneReplica on our local system.
We evaluated the architecture, data flow, and coordination mechanisms used by SceneReplica, including the services and nodes per stage, the parameterization of components, publish/subscribe schemes, and runtime control logic. 
The modular design of SceneReplica aligns with COMPARE's emphasis on separable and replaceable components in manipulation pipelines, also mirroring our goal of enabling plug-and-play evaluation across pipelines.
With SceneReplica serving as a reference implementation for vision-based manipulation, the strategies it employs can also inform the development of \textbf{pipeline-level} standards. 

\subsection{Developing New Modular Pipelines}

For COMPARE, we aim to achieve levels of modularity similar to efforts like SceneReplica, so we are gleaning best practices while also advancing the approach.
Our vision is to develop infrastructure tools in ROS  2 that can be used by the community to create pipelines that control the flow of an experiment wherein users utilize existing ROS packages or create their own components that follow the established standards and guidelines to ``drop in.''
Components should be interchangeable to be readily swapped out without having to write new code and the user should make minimal modifications to the infrastructure set-up to match their testing needs.
The infrastructure should also be sufficiently hardware-agnostic, assuming that the hardware is ROS compatible and a ROS driver exists; see the PickNik ROS  2 Compatible Hardware list\footnote{https://picknik.ai/hardware-ecosystem/}.
However, accommodations for ROS  1 must also be made as many open-source contributions are not yet available for ROS  2.
Prior work to develop this infrastructure is reviewed in \cite{flynn2025developing}.

\begin{figure}[htp]
    \centering
\begin{tikzpicture}[node distance=4mm and -12mm, every node/.style={align=center}]

\node[draw, rounded corners, minimum width=37.5mm, fill=lightgray!20, xshift=-21mm,font=\small] (perception_states) {States \\[65mm]};

\node[draw, rounded corners, minimum width=35mm, minimum height=4.25mm, fill=yellow!30, anchor=north, yshift=-6.5mm] at (perception_states.north) (get_point_cloud_service_state) {\scriptsize GetPointCloud \\[2mm]};
\node[draw, rectangle, minimum height=4.25mm, fill=blue!10, xshift=-8.75mm, yshift=-6.5mm] at (get_point_cloud_service_state.north) (get_point_cloud_camera_topic) {\scriptsize camera\_topic};
\node[draw, rectangle, minimum height=4.25mm, right= 0.5mm of get_point_cloud_camera_topic, fill=green!10] (get_point_cloud_full_cloud) {\scriptsize full\_cloud};
\node[draw, rectangle, minimum height=4.25mm, right= 0.5mm of get_point_cloud_full_cloud, fill=green!10] (get_point_cloud_etc) {\scriptsize ...};

\node[draw, rounded corners, minimum width=35mm, minimum height=4.25mm, below= 3.5mm of get_point_cloud_service_state, fill=yellow!30] (voxel_grid_filter_service_state) {\scriptsize VoxelGridFilter \\[2mm]};
\node[draw, rectangle, minimum height=4.25mm, fill=blue!10, xshift=-10.5mm, yshift=-6.5mm] at (voxel_grid_filter_service_state.north) (voxel_grid_filter_leaf_size) {\scriptsize leaf\_size};
\node[draw, rectangle, minimum height=4.25mm, right= 0.5mm of voxel_grid_filter_leaf_size, fill=green!10] (voxel_grid_filter_filtered_cloud) {\scriptsize filtered\_cloud};
\node[draw, rectangle, minimum height=4.25mm, right= 0.5mm of voxel_grid_filter_filtered_cloud, fill=green!10] (voxel_grid_filter_etc) {\scriptsize ...};

\draw[->] (get_point_cloud_service_state) -- (voxel_grid_filter_service_state); 

\node[draw, rounded corners, minimum width=35mm, minimum height=4.25mm, below= 3.5mm of voxel_grid_filter_service_state, fill=yellow!30] (outlier_removal_service_state) {\scriptsize OutlierRemoval \\[2mm]};
\node[draw, rectangle, minimum height=4.25mm, fill=blue!10, xshift=-10.5mm, yshift=-6.5mm] at (outlier_removal_service_state.north) (outlier_removal_radius) {\scriptsize radius\_max};
\node[draw, rectangle, minimum height=4.25mm, right= 0.5mm of outlier_removal_radius, fill=green!10] (outlier_removal_radius_filtered_cloud) {\scriptsize filtered\_cloud};
\node[draw, rectangle, minimum height=4.25mm, right= 0.5mm of outlier_removal_radius_filtered_cloud, fill=green!10] (outlier_removal_etc) {\scriptsize ...};

\draw[->] (voxel_grid_filter_service_state) -- (outlier_removal_service_state); 

\node[draw, rounded corners, minimum width=35mm, minimum height=4.25mm, below= 3.5mm of outlier_removal_service_state, fill=yellow!30] (plane_segmentation_service_state) {\scriptsize PlaneSegmentation \\[2mm]};
\node[draw, rectangle, minimum height=4.25mm, fill=green!10, xshift=-2.75mm, yshift=-6.5mm] at (plane_segmentation_service_state.north) (plane_segmentation_plane_indices) {\scriptsize plane\_indices};
\node[draw, rectangle, minimum height=4.25mm, right= 0.5mm of plane_segmentation_plane_indices, fill=green!10] (plane_segmentation_etc) {\scriptsize ...};

\draw[->] (outlier_removal_service_state) -- (plane_segmentation_service_state); 

\node[draw, rounded corners, minimum width=35mm, minimum height=4.25mm, below= 3.5mm of plane_segmentation_service_state, fill=yellow!30] (euclidean_cluster_extraction_service_state) {\scriptsize ClusterExtraction \\[2mm]};
\node[draw, rectangle, minimum height=4.25mm, fill=green!10, xshift=-2.75mm, yshift=-6.5mm] at (euclidean_cluster_extraction_service_state.north) (euclidean_cluster_extraction_cluster_indices) {\scriptsize cluster\_indices};
\node[draw, rectangle, minimum height=4.25mm, right= 0.5mm of euclidean_cluster_extraction_cluster_indices, fill=green!10] (euclidean_cluster_extraction_etc) {\scriptsize ...};

\draw[->] (plane_segmentation_service_state) -- (euclidean_cluster_extraction_service_state); 

\node[draw, rectangle, minimum width=40mm, right= 20.5mm of perception_states, fill=white!2, yshift=-6.25mm,font=\small] at (perception_states.north) (perception_input_keys) {Input Keys \\[4mm]};
\node[draw, rectangle, minimum height=4.25mm, fill=green!10, anchor=north, xshift=-11.5mm, yshift=-6.5mm] at (perception_input_keys.north) (input_key_full_cloud) {\scriptsize full\_cloud};
\node[draw, rectangle, minimum height=4.25mm, right= 1.5mm of input_key_full_cloud, fill=green!10] (input_key_filtered_cloud) {\scriptsize filtered\_cloud};
\node[draw, rectangle, minimum height=4.25mm, right= 1.5mm of input_key_filtered_cloud, fill=green!10] (input_keys_etc) {\scriptsize ...};

\node[draw, rectangle, minimum width=40mm, below= 1.5mm of perception_input_keys, fill=white!2,font=\small] (perception_output_keys) {Output Keys \\[10mm]};
\node[draw, rectangle, minimum height=4.25mm, fill=green!10, anchor=north, xshift=-11.5mm, yshift=-6.5mm] at (perception_output_keys.north) (output_key_full_cloud) {\scriptsize full\_cloud};
\node[draw, rectangle, minimum height=4.25mm, right= 1.5mm of output_key_full_cloud, fill=green!10] (output_key_cluster_sample_indices) {\scriptsize cluster\_indices};
\node[draw, rectangle, minimum height=4.25mm, below= 1.5mm of output_key_full_cloud, fill=green!10, xshift=7mm] (output_key_camera_pose) {\scriptsize camera\_pose};
\node[draw, rectangle, minimum height=4.25mm, right= 1.5mm of output_key_camera_pose, fill=green!10] (output_keys_etc) {\scriptsize ...};

\node[draw, rectangle, minimum width=40mm, below= 1.5mm of perception_output_keys, fill=white!2,font=\small] (perception_parameters_used) {Parameters Used \\[4mm]};
\node[draw, rectangle, minimum height=4.25mm, fill=blue!15, anchor=north, xshift=-9mm, yshift=-6.5mm] at (perception_parameters_used.north) (parameter_camera_topic) {\scriptsize camera\_topic};
\node[draw, rectangle, minimum height=4.25mm, right= 1.5mm of parameter_camera_topic, fill=blue!15] (parameters_used_leaf_size) {\scriptsize leaf\_size};
\node[draw, rectangle, minimum height=4.25mm, right= 1.5mm of parameters_used_leaf_size, fill=blue!15] (parameters_used_etc) {\scriptsize ...};

\begin{scope}[on background layer]              
    \node[draw=none, fill=none, inner sep=1.5mm, fit=(perception_input_keys) (perception_output_keys) (perception_states) (perception_parameters_used)] (perception_box) {} ; 
    \node[coordinate] (perception) at ($(perception_box.north)+(0,4mm)$) {};
\end{scope}

\begin{scope}[on background layer]
  \node[draw=black, fill=pink!15, rounded corners, inner sep=0pt, fit=(perception_box)(perception)] (perception_box_shader) {};       
  \node[anchor=north, font=\scriptsize\bfseries, align=center, xshift=0mm, yshift=-1mm] at (perception_box_shader.north) {\textbf{Model-Free Perception}}; 
\end{scope}




\node[draw, rounded corners, minimum width=37.5mm, below= 13mm of perception_box, fill=lightgray!20, xshift=-21mm,font=\small] (grasping_states) {States \\[38mm]};

\node[draw, rounded corners, minimum width=35mm, minimum height=4.25mm, fill=yellow!30, anchor=north, yshift=-6.5mm] at (grasping_states.north) (detect_grasps_service_state) {\scriptsize DetectGrasps \\[2mm]};
\node[draw, rectangle, minimum height=4.25mm, fill=green!10, xshift=-11.125mm, yshift=-6.5mm] at (detect_grasps_service_state.north) (detect_grasps_full_cloud) {\scriptsize full\_cloud};
\node[draw, rectangle, minimum height=4.25mm, right= 0.5mm of detect_grasps_full_cloud, fill=green!10] (detect_grasps_cluster_indices) {\scriptsize cluster\_indices};
\node[draw, rectangle, minimum height=4.25mm, right= 0.5mm of detect_grasps_cluster_indices, fill=green!10] (detect_grasps_etc) {\scriptsize ...};

\node[draw, rounded corners, minimum width=35mm, minimum height=4.25mm, below= 3.5mm of detect_grasps_service_state, fill=yellow!30] (filter_candidates_service_state) {\scriptsize FilterCandidates \\[2mm]};
\node[draw, rectangle, minimum height=4.25mm, fill=green!10, xshift=-2.25mm, yshift=-6.5mm] at (filter_candidates_service_state.north) (filter_candidates_list) {\scriptsize grasp\_candidates};
\node[draw, rectangle, minimum height=4.25mm, right= 0.5mm of filter_candidates_list, fill=green!10] (filter_candidates_etc) {\scriptsize ...};

\draw[->] (detect_grasps_service_state) -- (filter_candidates_service_state); 

\node[draw, rounded corners, minimum width=35mm, minimum height=4.25mm, below= 3.5mm of filter_candidates_service_state, fill=yellow!30] (calculate_grasp_poses_service_state) {\scriptsize CalculateGraspPoses \\[2mm]};
\node[draw, rectangle, minimum height=4.25mm, fill=blue!10, xshift=-9.5mm, yshift=-6.5mm] at (calculate_grasp_poses_service_state.north) (calculate_grasp_poses_gripper_width) {\scriptsize tool\_width};
\node[draw, rectangle, minimum height=4.25mm, right= 0.5mm of calculate_grasp_poses_gripper_width, fill=green!10] (calculate_grasp_poses_grasp_candidates) {\scriptsize grasp\_poses};
\node[draw, rectangle, minimum height=4.25mm, right= 0.5mm of calculate_grasp_poses_grasp_candidates, fill=green!10] (calculate_grasp_poses_etc) {\scriptsize ...};

\draw[->] (filter_candidates_service_state) -- (calculate_grasp_poses_service_state); 

\node[draw, rectangle, minimum width=40mm, right= 20.5mm of grasping_states, fill=white!2, yshift=-6.25mm,font=\small] at (grasping_states.north) (grasping_input_keys) {Input Keys \\[4mm]};
\node[draw, rectangle, minimum height=4.25mm, fill=green!10, anchor=north, xshift=-12mm, yshift=-6.5mm] at (grasping_input_keys.north) (grasping_input_key_full_cloud) {\scriptsize full\_cloud};
\node[draw, rectangle, minimum height=4.25mm, right= 1.5mm of grasping_input_key_full_cloud, fill=green!10] (grasping_input_key_cluster_indices) {\scriptsize cluster\_indices};
\node[draw, rectangle, minimum height=4.25mm, right= 1.5mm of grasping_input_key_cluster_indices, fill=green!10] (input_keys_etc) {\scriptsize ...};

\node[draw, rectangle, minimum width=40mm, below= 1.5mm of grasping_input_keys, fill=white!2,font=\small] (grasping_output_keys) {Output Keys \\[4mm]};
\node[draw, rectangle, minimum height=4.25mm, fill=green!10, anchor=north, xshift=-3.5mm, yshift=-6.5mm] at (grasping_output_keys.north) (grasping_output_key_grasp_poses) {\scriptsize grasp\_poses};
\node[draw, rectangle, minimum height=4.25mm, right= 1.5mm of grasping_output_key_grasp_poses, fill=green!10] (output_keys_etc) {\scriptsize ...};

\node[draw, rectangle, minimum width=40mm, below= 1.5mm of grasping_output_keys, fill=white!2,font=\small] (grasping_parameters_used) {Parameters Used \\[4mm]};
\node[draw, rectangle, minimum height=4.25mm, fill=blue!15, anchor=north, xshift=-3.5mm, yshift=-6.5mm] at (grasping_parameters_used.north) (grasping_parameter_used_tool_width) {\scriptsize tool\_width};
\node[draw, rectangle, minimum height=4.25mm, right= 1.5mm of grasping_parameter_used_tool_width, fill=blue!15] (grasping_parameters_used_etc) {\scriptsize ...};

\begin{scope}[on background layer]              
    \node[draw=none, fill=none, inner sep=1.5mm, fit= (grasping_states) (grasping_input_keys) (grasping_output_keys) (grasping_parameters_used)] (grasping_box) {} ; 
    \node[coordinate] (grasping) at ($(grasping_box.north)+(0,4mm)$) {};
\end{scope}

\begin{scope}[on background layer]
  \node[draw=black, fill=pink!15, rounded corners, inner sep=0pt, fit=(grasping_box) (grasping)] (grasping_box_shader) {};       
  \node[anchor=north, font=\scriptsize\bfseries, align=center, xshift=0mm, yshift=-1mm] at (grasping_box_shader.north) {\textbf{Model-Free Grasping (GPD)}}; 
\end{scope}

\draw[->] (perception_box_shader) -- (grasping_box_shader);



\node[draw, rounded corners, minimum width=37.5mm, below= 13mm of grasping_box, fill=lightgray!20, xshift=-21mm,font=\small] (pick_and_place_states) {States \\[38mm]};

\node[draw, rounded corners, minimum width=35mm, minimum height=4.25mm, fill=yellow!30, anchor=north, yshift=-6.5mm] at (pick_and_place_states.north) (move_to_approach_service_state) {\scriptsize MoveToApproach \\[2mm]};
\node[draw, rectangle, minimum height=4.25mm, fill=green!10, xshift=-2.5mm, yshift=-6.5mm] at (move_to_approach_service_state.north) (move_to_approach_grasp_poses) {\scriptsize grasp\_poses};
\node[draw, rectangle, minimum height=4.25mm, right= 0.5mm of move_to_approach_grasp_poses, fill=green!10] (detect_grasps_etc) {\scriptsize ...};

\node[draw, rounded corners, minimum width=35mm, minimum height=4.25mm, below= 3.5mm of move_to_approach_service_state, fill=yellow!30] (perform_grasp_service_state) {\scriptsize PerformGrasp \\[2mm]};
\node[draw, rectangle, minimum height=4.25mm, fill=green!10, xshift=-11.5mm, yshift=-6.5mm] at (perform_grasp_service_state.north) (perform_grasp_grasp_poses) {\scriptsize grasp\_poses};
\node[draw, rectangle, minimum height=4.25mm, right= 0.5mm of perform_grasp_grasp_poses, fill=green!10] (perform_grasp_outcome) {\scriptsize outcome};
\node[draw, rectangle, minimum height=4.25mm, right= 0.5mm of perform_grasp_outcome, fill=blue!10] (perform_grasp_step_size) {\scriptsize step\_size};

\draw[->] (move_to_approach_service_state) -- (perform_grasp_service_state); 

\node[draw, rounded corners, minimum width=35mm, minimum height=4.25mm, below= 3.5mm of perform_grasp_service_state, fill=yellow!30] (dropoff_service_state) {\scriptsize MoveToDropoff \\[2mm]};
\node[draw, rectangle, minimum height=4.25mm, fill=blue!10, xshift=-3mm, yshift=-6.5mm] at (dropoff_service_state.north) (dropoff_dropoff_pose) {\scriptsize dropoff\_pose};
\node[draw, rectangle, minimum height=4.25mm, right= 0.5mm of dropoff_dropoff_pose, fill=green!10] (dropoff_etc) {\scriptsize ...};

\draw[->] (perform_grasp_service_state) -- (dropoff_service_state); 

\node[draw, rectangle, minimum width=40mm, right= 20.5mm of pick_and_place_states, fill=white!2, yshift=-6.25mm,font=\small] at (pick_and_place_states.north) (pick_and_place_input_keys) {Input Keys \\[4mm]};
\node[draw, rectangle, minimum height=4.25mm, fill=green!10, anchor=north, xshift=-3mm, yshift=-6.5mm] at (pick_and_place_input_keys.north) (pick_and_place_grasp_poses) {\scriptsize grasp\_poses};
\node[draw, rectangle, minimum height=4.25mm, right= 1.5mm of pick_and_place_grasp_poses, fill=green!10] (input_keys_etc) {\scriptsize ...};

\node[draw, rectangle, minimum width=40mm, below= 1.5mm of pick_and_place_input_keys, fill=white!2,font=\small] (pick_and_place_output_keys) {Output Keys \\[4mm]};
\node[draw, rectangle, minimum height=4.25mm, fill=green!10, anchor=north, xshift=-3mm, yshift=-6.5mm] at (pick_and_place_output_keys.north) (pick_and_place_success_bool) {\scriptsize outcome};
\node[draw, rectangle, minimum height=4.25mm, right= 0.75mm of pick_and_place_success_bool, fill=green!10] (output_keys_etc) {\scriptsize ...};

\node[draw, rectangle, minimum width=40mm, below= 1.5mm of pick_and_place_output_keys, fill=white!2,font=\small] (pick_and_place_parameters_used) {Parameters Used \\[4mm]};
\node[draw, rectangle, minimum height=4.25mm, fill=blue!15, anchor=north, xshift=-11.5mm, yshift=-6.5mm] at (pick_and_place_parameters_used.north) (pick_and_place_parameters_used_cartesian_params) {\scriptsize step\_size};
\node[draw, rectangle, minimum height=4.25mm, right= 1.5mm of pick_and_place_parameters_used_cartesian_params, fill=blue!15] (pick_and_place_parameters_used_dropoff_pose) {\scriptsize dropoff\_pose};
\node[draw, rectangle, minimum height=4.25mm, right= 1.5mm of pick_and_place_parameters_used_dropoff_pose, fill=blue!15] (pick_and_place_parameters_used_etc) {\scriptsize ...};

\begin{scope}[on background layer]              
    \node[draw=none, fill=none, inner sep=1.5mm, fit= (pick_and_place_states) (pick_and_place_input_keys) (pick_and_place_output_keys) (pick_and_place_parameters_used)] (pick_and_place_box) {} ; 
    \node[coordinate] (pick_and_place) at ($(pick_and_place_box.north)+(0,4mm)$) {};
\end{scope}

\begin{scope}[on background layer]
  \node[draw=black, fill=pink!15, rounded corners, inner sep=0pt, fit=(pick_and_place_box) (pick_and_place)] (pick_and_place_box_shader) {};       
  \node[anchor=north, font=\scriptsize\bfseries, align=center, xshift=0mm, yshift=-1mm] at (pick_and_place_box_shader.north) {\textbf{Pick and Place Task}}; 
\end{scope}

\draw[->] (grasping_box_shader) -- (pick_and_place_box_shader);




\end{tikzpicture}
    \caption{Block diagram for perception-to-action pipelines in FlexBE using ROS  2 with state-based function calling and control indicating examples of different data types that are accessed and modified.}
    \label{fig:FlexBE_flow}
\end{figure}
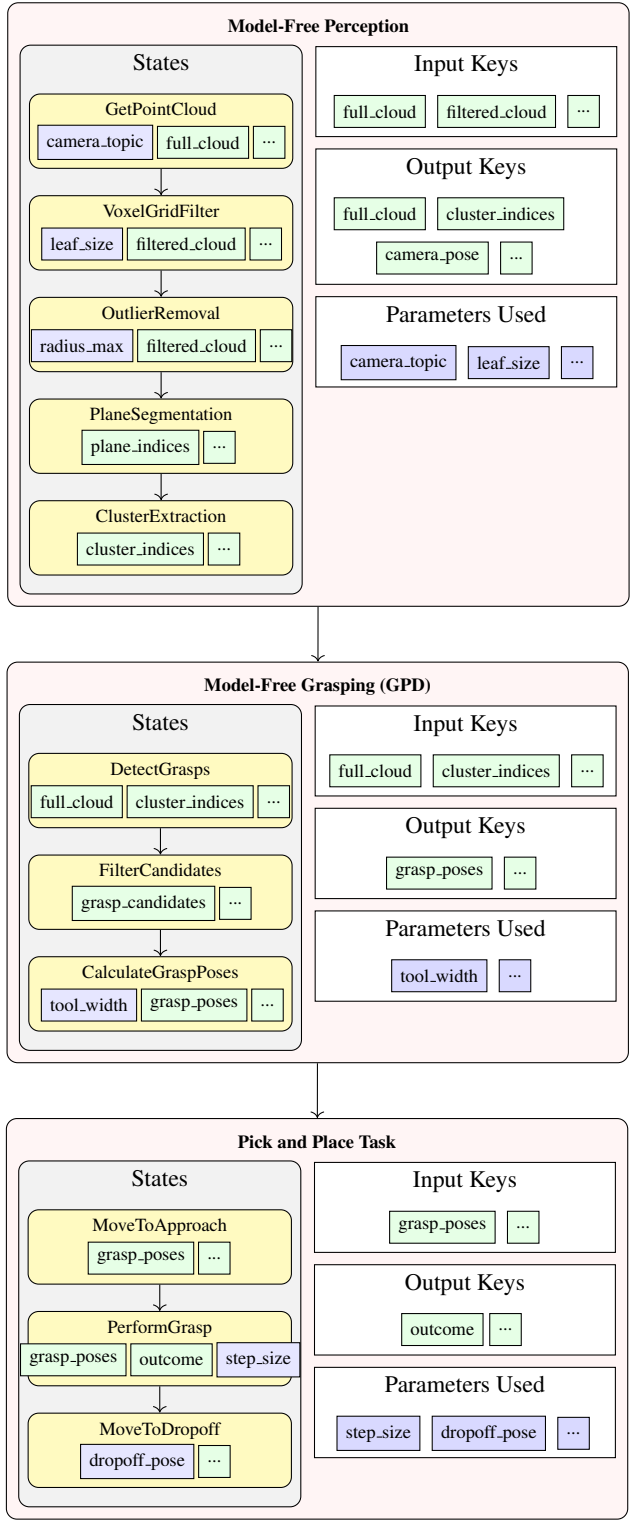

To this end, we are developing new pipelines that rely on FlexBE, a high-level behavior engine for ROS that is ``designed to ease the development and execution of complex robotic behaviors through the use of Hierarchical Finite State Machines (HFSMs).''~\footnote{https://flexbe.readthedocs.io/en/latest/}
See Fig.~\ref{fig:FlexBE_flow} for a block diagram of an example perception-to-action pipeline using FlexBE to perform \textit{model-free grasping} using Grasp Pose Detection (GPD)~\cite{ten2017grasp}.
The diagram in Fig.~\ref{fig:FlexBE_flow} reflects our novel modular pipeline implemented for ROS  2, which integrates state-based function calls with flexible control logic.
The FlexBE states are constructed such that each requests a single service and ultimately performs a single task to allow for a high degree of modularity when building a behavior. 
FlexBE also provides a drag-and-drop interface, so any states (or group of states) within a behavior can be modified on the fly to add, remove, or the change the order of states as needed. 
It is also trivial to add in timers or additional checks between states to assist with debugging during development or towards evaluating performance metrics as part of a benchmarking protocol. 
Each state is constructed such that it can be templated, with ample comments and boilerplate code to ensure that new states are simple and straightforward to develop.
Creating a new service state requires knowledge of the service topic name, the service request and response fields, and some knowledge of how user data and parameters are utilized within FlexBE.




User data variables and their values are handled within the FlexBE behavior and act as shared resources that can be accessed by any state as long as they are defined as \textit{input keys} or \textit{output keys} in the state code. 
Parameters in FlexBE function very similarly to ROS parameters; they are shared and accessible, but not modifiable during runtime.
Because of this, no data is truly ``passed'' between states, but each state accesses the live data during runtime and then uses that data to make a request to a service or action. 
Once that service or action returns a response, there is another opportunity for the state to update the internal user data before the state completes and the next state is allowed to start based on the triggered outcome. 
Parameters are typically only modified at the beginning of an experiment and are meant to set conditions; for example, setting a max counter value as part of evaluating a performance metric. 
Essentially, these values can be used to access persistent measurements or system conditions the user wants to protect.

Similar to states, service server and action server nodes are constructed with a single goal in order to promote and exercise modularity. 
Each node waits for a request, performs a specific function such as sending MoveIt commands to a robot, processing a point cloud, or even stopping or starting a recording software for well-timed data capture.
This structure is also fairly templated and the structure for creating a node is maintained across different nodes with regards to ROS  2 functionality and object-oriented structure.

The primary computer executing the pipeline contains the FlexBE components including states and behaviors, as well as the perception, planning, and execution components which are aided by compartmentalized utilities. 
Utilities come in the form of services and action servers where appropriate -- nodes in the background which perform an operation when requested and return a result often containing processed information and outcomes. 
To allow ROS 1 packages to be leveraged (e.g., GPD\footnote{https://github.com/atenpas/gpd}), we use a second machine consisting of a ROS  1 and ROS  2 workspace such that we can use the ros1\_bridge\footnote{\url{https://github.com/ROS2/ros1_bridge/}} to translate between the two environments across a network. 
This allows existing resources to be used without needing to migrate code before integration.

Our work-in-progress repository for all FlexBE assets can be found on GitHub\footnote{\url{https://github.com/uml-robotics/compare_flexbe}}; we intend to release the infrastructure in a Docker container to further streamline reproducibility.

\section{Next Steps and Conclusion}

In the near term, we will use our developed infrastructure to produce pipelines (like that shown in Fig.~\ref{fig:FlexBE_flow}) to integrate a variety of open-source components and conduct extensive side-by-side benchmarking for pick-and-place -- following the SceneReplica protocol --  and develop draft specifications of the proposed standards and guidelines that are used to drive the pipelines.
The contributions of this exercise are three-fold: 
\begin{enumerate}
    \item A set of draft guidelines for component-level open-source software components and standards for pipeline-level open-source software pipelines,
    \item Several example functional implementations of robot manipulation pipelines that leverage the standards and guidelines, and 
    \item An extensive set of benchmarks of truly comparable modular pipelines with a variety of components.
\end{enumerate}

For now, we expect to hold the motion planning components static (e.g., OMPL in MoveIt) while varying perception and grasp planning.
Several grasp planners (in addition to the previously mentioned 6-DoF GraspNet, Contact-GraspNet, GraspIt!, and GPD) have already been identified as potential options for this exercise, including 
DeepRLManip~\cite{lundell2021ddgc}
Grasp Proposal Network (GP-Net)~\cite{konrad2023gp}, 
GraspSAM~\cite{gualtieri2018learning}, 
ICG-Net~\cite{zurbrugg2024icgnet},  
MISCGrasp~\cite{fan2025miscgrasp}, 
PointNetGPD~\cite{liang2019pointnetgpd}, 
UniGrasp~\cite{shao2020unigrasp}, and
Volumetric Grasp Network (VGN)~\cite{breyer2021volumetric}. 
Table \ref{tab:grasp_planner} shows the relevant characteristics of each of these grasp planners for integration into perception-to-action pipelines: the particular input data each utilizes to plan grasps, the format of their output poses, and the camera point of view used for data collection.
We expect to develop different classes of \textbf{pipeline-level} standards corresponding to these characteristics.
For example: ``all grasp planners that output a 6-DoF grasp pose shall do so using X specified format as a standard input to the motion planner in the pipeline.''
Following such a standard would streamline swapping out the grasp planner in a pipeline to more easily compare performance.

\begin{table}[h!]
\small
\centering
\caption{Grasp planners and their relevant characteristics for integration into perception-to-action pipelines.}
\label{tab:grasp_planner}

\begin{tabular}{|l|l|l|l|}
\hline
\textbf{\begin{tabular}[c]{@{}l@{}}Grasp \\ Planner\end{tabular}} & \textbf{Input Data} & \textbf{Output Pose} & \textbf{\begin{tabular}[c]{@{}l@{}}Camera \\ PoV\end{tabular}} \\ \hline
\begin{tabular}[c]{@{}l@{}}6-DoF \\ GraspNet\end{tabular} & Point cloud & 6-DoF grasp pose & \begin{tabular}[c]{@{}l@{}}Eye-in-\\ hand\end{tabular} \\ \hline
\begin{tabular}[c]{@{}l@{}}Contact-\\ GraspNet\end{tabular} & Point cloud & 6-DoF grasp pose & \begin{tabular}[c]{@{}l@{}}Eye-in-\\ hand\end{tabular} \\ \hline
\begin{tabular}[c]{@{}l@{}}DeepRL-\\ Manip\end{tabular} & Depth image & Grasp policy & \begin{tabular}[c]{@{}l@{}}Eye-in-\\ hand\end{tabular} \\ \hline
GP-Net & Point cloud & 6-DoF grasp pose & Agnostic \\ \hline
GPD & Point cloud & 6-DoF grasp pose & \begin{tabular}[c]{@{}l@{}}Eye-in-\\ hand\end{tabular} \\ \hline
GraspIt! & \begin{tabular}[c]{@{}l@{}}3D object \\ model mesh\end{tabular} & \begin{tabular}[c]{@{}l@{}}6-DoF grasp pose \\ and force closure \\ metrics\end{tabular} & \begin{tabular}[c]{@{}l@{}}Eye-in-\\ hand\end{tabular} \\ \hline
GraspSAM & RGB image & \begin{tabular}[c]{@{}l@{}}2D grasp \\ rectangle\end{tabular} & Overhead \\ \hline
ICG-Net & Point cloud & 6-DoF grasp pose & Oblique \\ \hline
\begin{tabular}[c]{@{}l@{}}MISC\\ Grasp\end{tabular} & Depth image & 6-DoF grasp pose & \begin{tabular}[c]{@{}l@{}}Eye-in-\\ hand\end{tabular} \\ \hline
\begin{tabular}[c]{@{}l@{}}PointNet\\ GPD\end{tabular} & Point cloud & 6-DoF grasp pose & \begin{tabular}[c]{@{}l@{}}Multi-\\ view\end{tabular} \\ \hline
UniGrasp & Point cloud & Grasp points & \begin{tabular}[c]{@{}l@{}}Eye-in-\\ hand\end{tabular} \\ \hline
VGN & Depth image & 6-DoF grasp pose & \begin{tabular}[c]{@{}l@{}}Eye-in-\\ hand\end{tabular} \\ \hline
\end{tabular}

\end{table}

All materials will be iteratively shared with the COMPARE Ecosystem community to provide feedback.
We will encourage other institutions to replicate the example pipelines we provide to repeat the experiments in their own labs.
The leaderboard for the SceneReplica benchmarking protocol -- hosted on Robot-Manipulation.org\footnote{https://www.robot-manipulation.org/benchmarking/\\leaderboards/} -- will be updated to include the benchmarks derived from this exercise.
As participation grows, we hope to see others in the community adapt their open-source contributions to follow the COMPARE standards and guidelines to produce modular components that can be easily integrated into these pipelines.
Additionally, we intend to use the developed infrastructure to produce other types of pipelines such as those used for assembly and disassembly with the NIST Assembly Task Boards benchmarking effort~\cite{kimble2020benchmarking}.
We hope the community will similarly develop their own pipelines to further propagate the COMPARE vision.

We plan to establish a public leaderboard for benchmarking grasp planners using our new modular pipeline, following a protocol inspired by SceneReplica. 
Each selected planner will be tested locally on a Universal Robots UR5e arm with Robotiq 2F-85 end-effector and a wrist-mounted Intel RealSense.
While this camera point of view would be considered ``eye-in-hand'' by default, we will reposition the robot arm to capture snapshots as needed from multiple camera viewpoints providing input data as needed for each grasp planner. 
YCB objects will be arranged into cluttered experimental scenes, enabling consistent and reproducible evaluations. 
Performance will be quantified through success rates in both grasp execution and pick-and-place tasks, measured across physical trials in a near-to-far object order.

\section{Acknowledgments}

Thank you to the COMPARE Ecosystem community for their continued engagement towards improving open-source and benchmarking for robot manipulation.
As a community-based effort, the COMPARE Ecosystem relies on participation from robot manipulation researchers, open-source developers, and industry users.
All interested parties are encouraged to provide feedback by visiting Robot-Manipulation.org, join the Slack and/or Google Group, and reach out to the organizers.

Thank you to our collaborators that comprise the COMPARE project team: Kostas Bekris, Berk Calli, Aaron Dollar, Ricardo Digiovanni Frumento, Shambhuraj Mane, Daniel Nakhimovich, Vatsal Patel, Yu Sun, and Yifan Zhu.
This work and the COMPARE Ecosystem is supported in part by the National Science Foundation (TI-2346069).

\bibliography{references}

\begin{thebibliography}{30}
\providecommand{\natexlab}[1]{#1}

\bibitem[{Back et~al.(2025)Back, Lee, Kim, Rho, Lee, Kang, Lee, Noh, Lee, Lee et~al.}]{back2025graspclutter6d}
Back, S.; Lee, J.; Kim, K.; Rho, H.; Lee, G.; Kang, R.; Lee, S.; Noh, S.; Lee, Y.; Lee, T.; et~al. 2025.
\newblock GraspClutter6D: A Large-scale Real-world Dataset for Robust Perception and Grasping in Cluttered Scenes.
\newblock \emph{arXiv preprint arXiv:2504.06866}.

\bibitem[{Bottarel et~al.(2023)Bottarel, Altobelli, Pattacini, and Natale}]{bottarel2023graspa}
Bottarel, F.; Altobelli, A.; Pattacini, U.; and Natale, L. 2023.
\newblock GRASPA-fying the Panda: Easily Deployable, Fully Reproducible Benchmarking of Grasp Planning Algorithms.
\newblock \emph{IEEE Robotics \& Automation Magazine}.

\bibitem[{Breyer et~al.(2021)Breyer, Chung, Ott, Siegwart, and Nieto}]{breyer2021volumetric}
Breyer, M.; Chung, J.~J.; Ott, L.; Siegwart, R.; and Nieto, J. 2021.
\newblock Volumetric grasping network: Real-time 6 dof grasp detection in clutter.
\newblock In \emph{Conference on robot learning}, 1602--1611. PMLR.

\bibitem[{Calli et~al.(2015)Calli, Singh, Walsman, Srinivasa, Abbeel, and Dollar}]{calli2015ycb}
Calli, B.; Singh, A.; Walsman, A.; Srinivasa, S.; Abbeel, P.; and Dollar, A.~M. 2015.
\newblock The {YCB} object and model set: Towards common benchmarks for manipulation research.
\newblock In \emph{2015 International Conference on Advanced Robotics (ICAR)}, 510--517. IEEE.

\bibitem[{Cervera(2018)}]{cervera2018try}
Cervera, E. 2018.
\newblock Try to start it! the challenge of reusing code in robotics research.
\newblock \emph{IEEE robotics and automation letters}, 4(1): 49--56.

\bibitem[{Cervera(2023)}]{cervera2023run}
Cervera, E. 2023.
\newblock Run to the source: The effective reproducibility of robotics code repositories.
\newblock \emph{IEEE Robotics \& Automation Magazine}, 31(2): 125--134.

\bibitem[{Deng et~al.(2021)Deng, Mousavian, Xiang, Xia, Bretl, and Fox}]{deng2021poserbpf}
Deng, X.; Mousavian, A.; Xiang, Y.; Xia, F.; Bretl, T.; and Fox, D. 2021.
\newblock PoseRBPF: A Rao--Blackwellized particle filter for 6-D object pose tracking.
\newblock \emph{IEEE Transactions on Robotics}, 37(5): 1328--1342.

\bibitem[{Eng(2019)}]{eng2019qt5}
Eng, L.~Z. 2019.
\newblock \emph{Qt5 C++ GUI Programming Cookbook: Practical recipes for building cross-platform GUI applications, widgets, and animations with Qt 5}.
\newblock Packt Publishing Ltd.

\bibitem[{Fan et~al.(2025)Fan, Cai, Li, Jiao, Zheng, Lu, Liang, and Wang}]{fan2025miscgrasp}
Fan, Q.; Cai, Y.; Li, C.; Jiao, C.; Zheng, X.; Lu, T.; Liang, B.; and Wang, S. 2025.
\newblock MISCGrasp: Leveraging Multiple Integrated Scales and Contrastive Learning for Enhanced Volumetric Grasping.
\newblock \emph{arXiv preprint arXiv:2507.02672}.

\bibitem[{Flynn et~al.(2025)Flynn, Bekris, Calli, Dollar, Norton, Sun, and Yanco}]{flynn2025developing}
Flynn, B.; Bekris, K.; Calli, B.; Dollar, A.; Norton, A.; Sun, Y.; and Yanco, H. 2025.
\newblock Developing modular grasping and manipulation pipeline infrastructure to streamline performance benchmarking.
\newblock \emph{arXiv preprint arXiv:2504.06819}.

\bibitem[{Gualtieri and Platt(2018)}]{gualtieri2018learning}
Gualtieri, M.; and Platt, R. 2018.
\newblock Learning 6-dof grasping and pick-place using attention focus.
\newblock In \emph{Conference on Robot Learning}, 477--486. PMLR.

\bibitem[{Hill(2000)}]{hill2000computer}
Hill, F.~J. 2000.
\newblock \emph{Computer graphics using OpenGL}.
\newblock Prentice Hall PTR.

\bibitem[{Khargonkar et~al.(2024)Khargonkar, Allu, Lu, Prabhakaran, Xiang et~al.}]{khargonkar2024scenereplica}
Khargonkar, N.; Allu, S.~H.; Lu, Y.; Prabhakaran, B.; Xiang, Y.; et~al. 2024.
\newblock Scenereplica: Benchmarking real-world robot manipulation by creating replicable scenes.
\newblock In \emph{2024 IEEE International Conference on Robotics and Automation (ICRA)}, 8258--8264. IEEE.

\bibitem[{Kimble et~al.(2020)Kimble, Van~Wyk, Falco, Messina, Sun, Shibata, Uemura, and Yokokohji}]{kimble2020benchmarking}
Kimble, K.; Van~Wyk, K.; Falco, J.; Messina, E.; Sun, Y.; Shibata, M.; Uemura, W.; and Yokokohji, Y. 2020.
\newblock Benchmarking protocols for evaluating small parts robotic assembly systems.
\newblock \emph{IEEE Robotics and Automation Letters}, 5(2): 883--889.

\bibitem[{Konrad, McDonald, and Villing(2023)}]{konrad2023gp}
Konrad, A.; McDonald, J.; and Villing, R. 2023.
\newblock GP-net: Flexible Viewpoint Grasp Proposal.
\newblock In \emph{2023 21st International Conference on Advanced Robotics (ICAR)}, 317--324. IEEE.

\bibitem[{Liang et~al.(2019)Liang, Ma, Li, G{\"o}rner, Tang, Fang, Sun, and Zhang}]{liang2019pointnetgpd}
Liang, H.; Ma, X.; Li, S.; G{\"o}rner, M.; Tang, S.; Fang, B.; Sun, F.; and Zhang, J. 2019.
\newblock Pointnetgpd: Detecting grasp configurations from point sets.
\newblock In \emph{2019 International Conference on Robotics and Automation (ICRA)}, 3629--3635. IEEE.

\bibitem[{Lim et~al.(2024)Lim, Kim, Kim, Lee, and Park}]{lim2024equigraspflow}
Lim, B.; Kim, J.; Kim, J.; Lee, Y.; and Park, F.~C. 2024.
\newblock Equigraspflow: SE (3)-equivariant 6-dof grasp pose generative flows.
\newblock In \emph{8th Annual Conference on Robot Learning}.

\bibitem[{Lu et~al.(2022)Lu, Chen, Ruozzi, and Xiang}]{https://doi.org/10.48550/arxiv.2211.11679}
Lu, Y.; Chen, Y.; Ruozzi, N.; and Xiang, Y. 2022.
\newblock Mean Shift Mask Transformer for Unseen Object Instance Segmentation.

\bibitem[{Luebke(2008)}]{luebke2008cuda}
Luebke, D. 2008.
\newblock CUDA: Scalable parallel programming for high-performance scientific computing.
\newblock In \emph{2008 5th IEEE international symposium on biomedical imaging: from nano to macro}, 836--838. IEEE.

\bibitem[{Lundell, Verdoja, and Kyrki(2021)}]{lundell2021ddgc}
Lundell, J.; Verdoja, F.; and Kyrki, V. 2021.
\newblock Ddgc: Generative deep dexterous grasping in clutter.
\newblock \emph{IEEE Robotics and Automation Letters}, 6(4): 6899--6906.

\bibitem[{Miller and Allen(2004)}]{miller2004graspit}
Miller, A.~T.; and Allen, P.~K. 2004.
\newblock Graspit! a versatile simulator for robotic grasping.
\newblock \emph{IEEE Robotics \& Automation Magazine}, 11(4): 110--122.

\bibitem[{Mousavian, Eppner, and Fox(2019)}]{mousavian20196}
Mousavian, A.; Eppner, C.; and Fox, D. 2019.
\newblock 6-dof graspnet: Variational grasp generation for object manipulation.
\newblock In \emph{Proceedings of the IEEE/CVF International Conference on Computer Vision}, 2901--2910.

\bibitem[{Quigley et~al.(2009)Quigley, Conley, Gerkey, Faust, Foote, Leibs, Berger, Wheeler, and Ng}]{quigley2009ros}
Quigley, M.; Conley, K.; Gerkey, B.; Faust, J.; Foote, T.; Leibs, J.; Berger, E.; Wheeler, R.; and Ng, A. 2009.
\newblock {ROS}: an open-source Robot Operating System.
\newblock In \emph{ICRA Workshop on Open Source Software}, volume~3, 5. Kobe, Japan.

\bibitem[{Schroeder, Avila, and Hoffman(2000)}]{schroeder2000visualizing}
Schroeder, W.~J.; Avila, L.~S.; and Hoffman, W. 2000.
\newblock Visualizing with VTK: a tutorial.
\newblock \emph{IEEE Computer graphics and applications}, 20(5): 20--27.

\bibitem[{Shao et~al.(2020)Shao, Ferreira, Jorda, Nambiar, Luo, Solowjow, Ojea, Khatib, and Bohg}]{shao2020unigrasp}
Shao, L.; Ferreira, F.; Jorda, M.; Nambiar, V.; Luo, J.; Solowjow, E.; Ojea, J.~A.; Khatib, O.; and Bohg, J. 2020.
\newblock Unigrasp: Learning a unified model to grasp with multifingered robotic hands.
\newblock \emph{IEEE Robotics and Automation Letters}, 5(2): 2286--2293.

\bibitem[{Sundermeyer et~al.(2021)Sundermeyer, Mousavian, Triebel, and Fox}]{sundermeyer2021contact}
Sundermeyer, M.; Mousavian, A.; Triebel, R.; and Fox, D. 2021.
\newblock Contact-graspnet: Efficient 6-dof grasp generation in cluttered scenes.
\newblock In \emph{2021 IEEE International Conference on Robotics and Automation (ICRA)}, 13438--13444. IEEE.

\bibitem[{Ten~Pas et~al.(2017)Ten~Pas, Gualtieri, Saenko, and Platt}]{ten2017grasp}
Ten~Pas, A.; Gualtieri, M.; Saenko, K.; and Platt, R. 2017.
\newblock Grasp pose detection in point clouds.
\newblock \emph{The International Journal of Robotics Research}, 36(13-14): 1455--1473.

\bibitem[{Xiang et~al.(2017)Xiang, Schmidt, Narayanan, and Fox}]{xiang2017posecnn}
Xiang, Y.; Schmidt, T.; Narayanan, V.; and Fox, D. 2017.
\newblock Posecnn: A convolutional neural network for 6d object pose estimation in cluttered scenes.
\newblock \emph{arXiv preprint arXiv:1711.00199}.

\bibitem[{Xiang et~al.(2020)Xiang, Xie, Mousavian, and Fox}]{xiang2020learning}
Xiang, Y.; Xie, C.; Mousavian, A.; and Fox, D. 2020.
\newblock Learning RGB-D Feature Embeddings for Unseen Object Instance Segmentation.
\newblock In \emph{Conference on Robot Learning (CoRL)}.

\bibitem[{Zurbr{\"u}gg et~al.(2024)Zurbr{\"u}gg, Liu, Engelmann, Kumar, Hutter, Patil, and Yu}]{zurbrugg2024icgnet}
Zurbr{\"u}gg, R.; Liu, Y.; Engelmann, F.; Kumar, S.; Hutter, M.; Patil, V.; and Yu, F. 2024.
\newblock ICGNet: a unified approach for instance-centric grasping.
\newblock In \emph{2024 IEEE International Conference on Robotics and Automation (ICRA)}, 4140--4146. IEEE.

\end{thebibliography}

\end{document}